\documentclass[letterpaper, 10 pt, journal, twoside]{IEEEtran}

\usepackage{amsmath,amsfonts}
\usepackage{algorithmic}
\usepackage{array}
\usepackage[caption=false,font=normalsize,labelfont=sf,textfont=sf]{subfig}
\usepackage{textcomp}
\usepackage{stfloats}
\usepackage{url}
\usepackage{verbatim}
\usepackage{graphicx}
\def\BibTeX{{\rm B\kern-.05em{\sc i\kern-.025em b}\kern-.08em
    T\kern-.1667em\lower.7ex\hbox{E}\kern-.125emX}}
\usepackage{balance}

\usepackage{comment}
\usepackage{xcolor}
\usepackage{multirow}
\usepackage{times}
\usepackage{epsfig}
\usepackage{amsmath}
\usepackage{amssymb}
\usepackage{bm}
\usepackage{siunitx}
\usepackage{graphics} 
\usepackage{graphicx}
\usepackage{float}
\usepackage{mathrsfs}
\usepackage{booktabs}
\usepackage{cite}
\usepackage{helvet}
\usepackage{fontawesome5}

\definecolor{lightviolet}{rgb}{0.722, 0.549, 0.867}
\definecolor{lightred}{rgb}{1.0, 0.388, 0.388}
\usepackage[pagebackref=false,breaklinks=true,colorlinks=true,bookmarks=false,linkcolor=lightred,citecolor=cyan,urlcolor=lightviolet]{hyperref}

\newcommand{\red}[1]{\textcolor{red}{#1}}

\newcommand{\black}[1]{\textcolor{black}{#1}}

\newcommand{\FMCWLIO}{%
\sffamily\bfseries
\textcolor[rgb]{0.945,0.149,0.114}{F}%
\textcolor[rgb]{1.000,0.573,0.000}{M}%
\textcolor[rgb]{0.992,0.859,0.165}{C}%
\textcolor[rgb]{0.302,0.894,0.169}{W}%
\textcolor[rgb]{0.059,0.965,0.573}{-}%
\textcolor[rgb]{0.043,0.847,0.976}{L}%
\textcolor[rgb]{0.173,0.133,0.984}{I}%
\textcolor[rgb]{0.765,0.176,0.953}{O}%
}

\newcommand{\FreeInit}{%
{\sffamily\bfseries\itshape\color[rgb]{0.95,0.35,0.35}Free-Init}%
}

\begin{document}


\title{
{\href{https://doi.org/10.1109/LRA.2024.3490395}{\FreeInit}} \\ {\vspace{5pt}}
Scan-Free, Motion-Free, and Correspondence-Free \\
Initialization for Doppler LiDAR-Inertial Systems
}

\author{Mingle Zhao, Jiahao Wang, Tianxiao Gao, Chengzhong Xu, and Hui Kong {\vspace{-8pt}}
\thanks{The authors are with the University of Macau.}
\thanks{\textbf{Publication}: M. Zhao, J. Wang, T. Gao, C. Xu and H. Kong, "Free-Init: Scan-Free, Motion-Free, and Correspondence-Free Initialization for Doppler LiDAR-Inertial Systems," in \textit{IEEE Robotics and Automation Letters}, vol. 9, no. 12, pp. 11329-11336, 2024, DOI: {\href{https://doi.org/10.1109/LRA.2024.3490395}{10.1109/LRA.2024.3490395}}.}
}

\maketitle



\begin{abstract}
Robust initialization is crucial for online systems. In the letter, a high-frequency and resilient initialization framework is designed for LiDAR-inertial systems, leveraging both inertial sensors and Doppler LiDAR. The innovative FMCW Doppler LiDAR opens up a novel avenue for robotic sensing by capturing not only point range but also Doppler velocity via the intrinsic Doppler effect. By fusing point-wise Doppler velocity with inertial measurements under non-inertial kinematics, the proposed framework, {\href{https://doi.org/10.1109/LRA.2024.3490395}{\FreeInit}}, eliminates reliance on motion undistortion of LiDAR scans, excitation motions, and map correspondences during the initialization phase. Free-Init is also plug-and-play compatible with typical LiDAR-inertial systems and is versatile to handle a wide range of initial motions when the system starts, including stationary, dynamic, and even violent motions. The embedded Doppler-inertial velocimeter ensures fast convergence and high-frequency performance, delivering outputs exceeding 10 kHz. Comprehensive experiments on diverse platforms and across myriad motion scenes validate the framework's effectiveness. The results demonstrate the superior performance of Free-Init, highlighting the necessity of fast, resilient, and dynamic initialization for online systems.
\end{abstract}




\phantomsection
\section*{Resources} \label{sec:resources}
\vspace{5pt}

    \begin{center}
    \begin{tabular}{@{}l@{\hspace{8pt}}c@{\hspace{8pt}}l@{}}
        IEEE Xplore Link & : & {\href{https://ieeexplore.ieee.org/document/10740796}{\FreeInit}} \\
        arXiv Paper Link & : & {\href{https://arxiv.org/abs/2609.29375}{{\red{\faFilePdf}}\enspace{Free-Init}}} \\
        Code \& Dataset & : & {\href{https://github.com/IMRL/Free-Init}{{\black{\faGithub}}{\enspace{Free-Init}}}} and {\href{https://github.com/IMRL/FMCW-LIO}{{\black{\faGithub}}{\enspace{FMCW-LIO}}}} \\
        Experiment Video & : & {\href{https://youtu.be/FbyzvJ-4bHI}{{\red{\faYoutube}}{\enspace{Free-Init}}}}
    \end{tabular}
    \end{center}


\section{Introduction} \label{sec:intro}

\IEEEPARstart{L}{ight} Detection And Ranging (LiDAR) sensors and Inertial Measurement Units (IMUs) are commonly employed to facilitate real-time state estimation and environmental mapping. LiDAR-inertial odometry (LIO) systems are widely applied in various field scenarios, including extreme environments \cite{lee2024lidar, ebadi2023present}. For mobile robots, online LIO not only provides state estimates for planners and controllers but also generates dense maps for perception and navigation. 

    \begin{figure}[!t]
       \centering
       \includegraphics[width=0.47\textwidth]{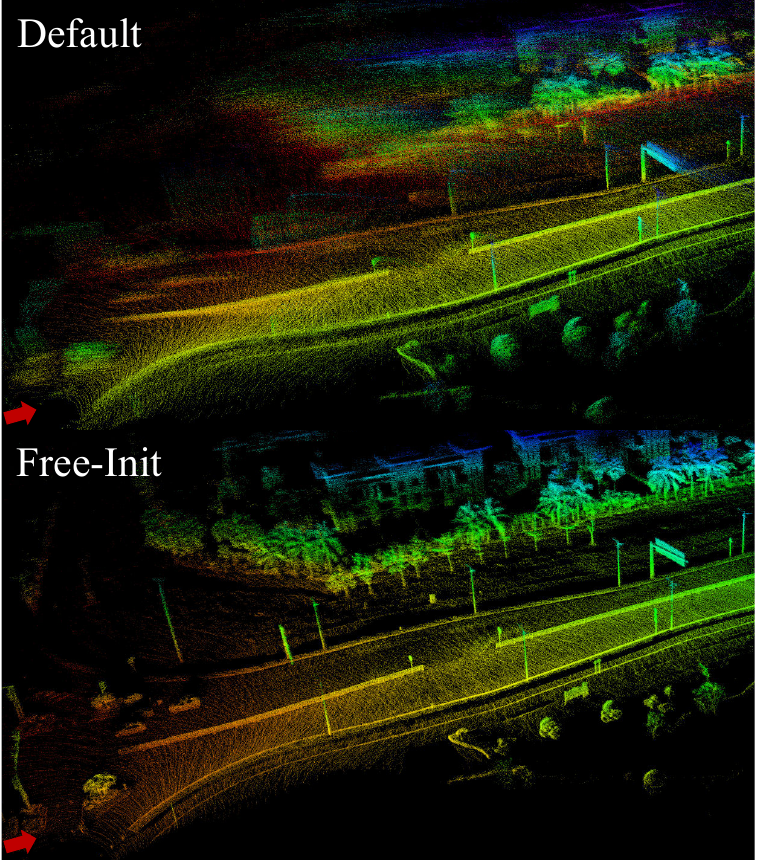}
       \caption{Mapping comparison with the default initialization and Free-Init when the platform is moving on the road. The red arrows indicate the platform positions and moving directions at the start of initialization. The rapid speed results in a highly blurred initial map generated with the default initialization. In contrast, the map generated with Free-Init is remarkably distinct, enabling sharp mapping of cars, poles, and traffic signs along the road.} 
       \label{fig:mapping_demo}
    \end{figure}

\subsection{Initialization in LiDAR-Inertial Odometry}
Owing to the intrinsic non-linearity, online LIO systems require accurate initial states (e.g., initial poses, velocities, IMU biases) or prior information (e.g., global poses) from the initialization module to ensure the accuracy, stability, and convergence of online estimators. However, there is a scarcity of research regarding the initialization in LIO systems. The initialization in mainstream LIO systems is straightforwardly based on the stationary motion assumption \cite{LINS, LIO-SAM, FAST-LIO, FAST-LIO2, DLIO}, where the initial rotation can be set to identity or be derived by normalization methods, the initial position and velocity are readily set to zero. Then IMU biases and the gravity vector can be computed using static IMU measurements. Some methods require additional sensors for initialization or calibration. In \cite{wang2021online}, the authors utilize a camera as an auxiliary sensor to calibrate the temporal-spatial offsets between LiDAR and IMU. Likewise, the Global Navigation Satellite System (GNSS) sensor is leveraged in \cite{taylor2015motion} to constrain the pose estimation. Besides, a minority of initialization methods rely on specific excitation motions to estimate temporal offsets, IMU biases, extrinsic parameters for subsequent LIO systems \cite{LI-Init}. 

However, despite the methods in \cite{LI-Init, wang2021online, taylor2015motion} can run online and serve as initialization modules before the LIO system starts, these methods necessitate additional sensors or specific excitation motions. Importantly, these methods essentially align with calibration methods rather than initialization methods. In addition, IMU biases, temporal-spatial offsets, and other states (e.g., poses and velocities) may vary upon each system startup. Moreover, it is challenging to perform excitation motions at each startup, particularly for specific configured platforms such as autonomous vehicles or large-scale drones. Essentially, following the design philosophy for rigorous and precise online systems, time synchronization and extrinsic calibration should be completed within the system design stage and prior to online deployment. Thus, the calibration of unknown parameters should not be handled within initialization modules but rather within calibration and identification processes. Notably, the true responsibility of initialization modules is to provide online estimators with reasonably accurate initial states and uncertainties, ensuring stability and convergence, irrespective of platform motions, operational modes, or environments, while preventing system divergences or failures.

\subsection{Crux and Ideal Initialization} \label{subsec:crux}
When the system starts, significant errors often manifest in the initial states. In LIO systems, these large initial errors can lead to cumulative errors in LiDAR scans and maps, which supply observations for state estimation through undistorted scans and map correspondences. Simultaneously, erroneous observations can further exacerbate estimation errors. This dependence between the estimator, LiDAR scans, and the online-established map, creates a positive feedback loop of errors, which can potentially lead to system divergence \cite{zhao2024FMCW-LIO}. Crucially, conventional LiDARs can only provide geometric observations which are directly associated with the known system poses and environmental structures, whereas initialization modules should inherently and independently provide accurate initial states to subsequent estimators, in the absence of known poses, velocities, and maps. Hence, employing conventional LiDAR data in initialization creates a similar chicken-and-egg paradox. Thereby, typical LIO systems \cite{LINS, FAST-LIO, FAST-LIO2, DLIO} rely solely on IMU data under certain motion assumptions during initialization. Similarly, dynamic initialization that relies on loosely-coupled LiDAR odometry \cite{LI-Init} fails to fundamentally address the above correspondence-dependency issue. Moreover, online initialization in degenerate scenes (e.g., tunnels, highways, flat terrains) can directly lead to system failures.  

Overall, an ideal initialization module for online LIO should be independent of platform motions, accumulated LiDAR scans, LiDAR maps, and environmental structures. Significantly, it should also be capable of offering accurate initial states and maps to the subsequent online estimator, regardless of \textit{when}, \textit{where}, and \textit{how} the system starts. 

Fortunately, rapid advancements in LiDAR sensors are unlocking new possibilities for robotic sensing. One notable development is the Frequency Modulated Continuous Wave (FMCW) Doppler LiDAR, which utilizes laser wave modulation in the frequency domain to capture instant range sensing and Doppler velocity \cite{DICP}. The Doppler measurements provide observations independent of poses and geometric structures \cite{DICP, Need_for_Speed, zhao2024FMCW-LIO}. As a result, adopting FMCW Doppler LiDARs sparks a novel route to designing the aforementioned ideal initialization framework and addresses the crux from the correspondence dependency inherent in conventional LIO systems. 

\subsection{Proposed Methodology} \label{subsec:proposed_method}
In this work, we propose Free-Init, an initialization framework for LIO systems. The framework overview is shown in Fig. \ref{fig:system_overview}. Free-Init takes LiDAR points and IMU data as inputs. The designed Doppler-inertial velocimeter estimates the LiDAR velocity, body velocity, and gyroscope bias under a point-wise scheme with extremely high-frequency outputs (over 10 kHz). An optimization problem is then established to estimate accelerometer bias and gravity. Finally, the estimated states, uncertainties, and static map are fed into the subsequent LIO as initial estimates. In \cite{Need_for_Speed}, the authors use the FMCW Doppler LiDAR to design a scan-based velocity estimator, achieving an open-loop, correspondence-free LIO within a continuous-time framework. However, this estimator is not suitable for full odometry and cannot achieve the same accuracy level as conventional LIO systems over long periods, as discussed in \cite{Need_for_Speed}. Contrarily, we design a discrete-time velocimeter under a point-wise updating manner without any interpolation models, enabling high-frequency performance for robust initialization. Meanwhile, Free-Init is a complete and resilient initialization framework, leveraging all FMCW Doppler LiDAR, gyroscope, and accelerometer data. 

In summary, the contributions are as follows:
\begin{enumerate}
    \item A complete, LiDAR scan-free, excitation motion-free, and map correspondence-free initialization framework, {\href{https://doi.org/10.1109/LRA.2024.3490395}{\FreeInit}}, is proposed for Doppler LiDAR-inertial systems. The framework design consistently aligns with the design philosophy of an ideal initialization module. 
    
    \item A novel high-frequency Doppler-inertial velocimeter is designed, exploiting the sensing nature of Doppler LiDARs with an efficient point-wise filtering scheme. 
    
    \item A formulation of Doppler observations, along with the estimation of accelerometer bias and gravity, is derived from a unified non-inertial kinematics perspective. 
    
    \item Extensively diverse experiments are conducted, demonstrating the effectiveness and robustness of Free-Init.
    
    \item The source code of Free-Init, the data sequences, the integration of Free-Init into the {\href{https://doi.org/10.1109/LRA.2024.3396636}{\FMCWLIO}} framework, and the experiment video are made publicly available (see the \nameref{sec:resources} section for details).
\end{enumerate}


\section{Notation and Preliminary} \label{sec:system_overview} 
\subsection{Notation}
We use $^{w}(\cdot)$, $^{b}(\cdot)$, and $^{l}(\cdot)$ to represent a 3D vector in the world, body, and LiDAR frame, respectively. The body frame $b$ coincides with the IMU frame, and the world frame $w$ is the first body frame $b_0$ when the system starts. The inertial frame is the world frame and the earth is static. The gravity ${^{w}\mathbf{g}}$ is constant in the world frame. For a 3D position $\mathbf{p} \in \mathbb{R}^3$ from point $A$ to point $B$ in the frame $C$ is denoted as $^{C}\mathbf{p}_{AB}$. A 3D velocity of $B$ with respect to $A$ expressed in $C$ is $^{C}\mathbf{v}_{AB}$. We use $\mathbf{R}_{AB} \! \in \! SO(3)$ to represent a rotation matrix rotating a vector expressed in the frame $B$ to $A$. $\mathbf{T}_{AB} \! \in \! SE(3)$ transforms a 3D vector in the frame $B$ to $A$. $\left[ \mathbf{t} \right]_{\times}$ denotes the $3\times 3$ skew-symmetric matrix of a 3D vector $\mathbf{t}$. $\mathbf{0}$ is the $3 \times 1$ zero vector, $\mathbf{0}_{n \times m}$ is the $n \times m$ zero matrix, $\mathbf{I}$ is the $3 \times 3$ identity matrix. 
$\widetilde{(\cdot)}$, $\widehat{(\cdot)}$, and $\overline{(\cdot)}$ respectively denote the measurement, the propagated state, and the updated state. 
\subsection{Non-Inertial Kinematics} \label{subsec:noninertial}
Considering the motion of a sensor frame $s$ with respect to the non-inertial body frame $b$, we use $^{b}\bm{\omega}_{wb}$ and $^{b}\mathbf{a}_{wb}$ to denote the angular velocity and the acceleration of body frame with respect to the world (inertial) frame $w$, respectively. Likewise, $^{s}\bm{\omega}_{ws}$ and $^{s}\mathbf{a}_{ws}$ denote the angular velocity and the acceleration of sensor frame with respect to the world frame. They are all expressed in their respective frames. Therefore, both first-order and second-order non-inertial kinematics can be derived \cite{landau1976mechanics}. For the translational and rotational parts, the first-order and second-order kinematics are as follows: 
    \begin{align}
        {{^b}\dot{\mathbf{p}}_{bs}} &= {{^b}\mathbf{v}_{ws}} - {{^b}\mathbf{v}_{wb}} - \left[ {^{b}\bm{\omega}_{wb}} \right]_{\times} {{^b}\mathbf{p}_{bs}} \label{eq:noninertial_kin_lin_1st} \\ 
        {{^b}\ddot{\mathbf{p}}_{bs}} &= {{^b}\mathbf{a}_{ws}} - {{^b}\mathbf{a}_{wb}} - {\left[ {^{b}\bm{\omega}_{wb}} \right]_{\times}^{2}} {{^b}\mathbf{p}_{bs}} - {\left[ {^{b}\dot{\bm{\omega}}_{wb}} \right]_{\times}} {{^b}\mathbf{p}_{bs}} \notag \\ 
        &\hspace{16pt} - 2 {\left[ {^{b}\bm{\omega}_{wb}} \right]_{\times}} {{^b}\dot{\mathbf{p}}_{bs}} \label{eq:noninertial_kin_lin_2nd} \\ 
        {\dot{\mathbf{R}}_{bs}} &= {\mathbf{R}_{bs}} {\left[ {^{s}\bm{\omega}_{ws}} - {\mathbf{R}_{bs}^{\top}} {^{b}\bm{\omega}_{wb}} \right]_{\times}} \label{eq:noninertial_kin_ang_1st} \\ 
        {\ddot{\mathbf{R}}_{bs}} &= {\mathbf{R}_{bs}} \left( {\left[ {^{s}\bm{\omega}_{ws}} - {\mathbf{R}_{bs}^{\top}} {^{b}\bm{\omega}_{wb}} \right]_{\times}^{2}} + {\left[ {^{s}\dot{\bm{\omega}}_{ws}} - {\mathbf{R}_{bs}^{\top}} {^{b}\dot{\bm{\omega}}_{wb}} \right]_{\times}} \right. \notag \\ 
        &\hspace{16pt} \left.  + {\left[ {\left( {^{s}\bm{\omega}_{ws}} - {\mathbf{R}_{bs}^{\top}} {^{b}\bm{\omega}_{wb}} \right)} \times {\left( {\mathbf{R}_{bs}^{\top}} {^{b}\bm{\omega}_{wb}} \right)} \right]_{\times}} \right) \label{eq:noninertial_kin_ang_2nd}
    \end{align}
    \begin{figure}[!t]
       \centering
       \includegraphics[width=0.47\textwidth]{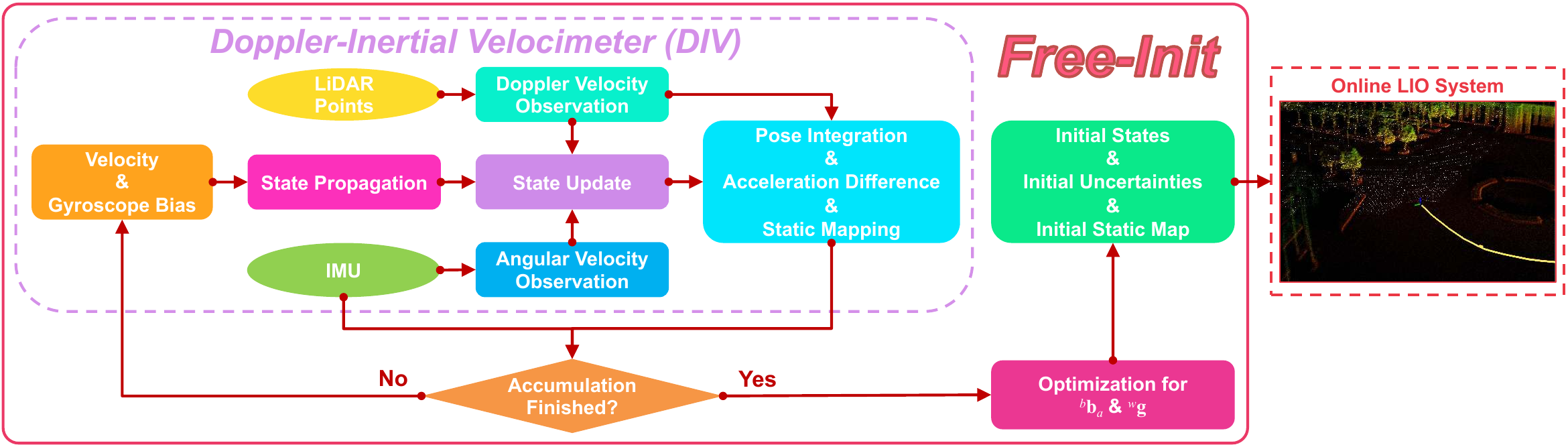}
       \caption{Framework overview. The initial states, uncertainties, and static map from Free-Init are fed into the subsequent LIO system.} 
       \label{fig:system_overview}
    \end{figure}

\section{Methodology} \label{sec:methodology}
\subsection{State and Problem Description}
\subsubsection{State in Typical LIO Systems} 
The state in general LIO systems usually includes the rotation and position of body frame with respect to the world frame, i.e., $\mathbf{R}_{wb}$ and $^{w}\mathbf{p}_{wb}$ (typically, the first body frame is denoted as the world frame), the body velocity in the world frame $^{w}\mathbf{v}_{wb}$, gyroscope and accelerometer biases $^{b}\mathbf{b}_{g}$ and $^{b}\mathbf{b}_{a}$ \cite{LINS, FAST-LIO, FAST-LIO2, LIO-SAM, DLIO}. In this work, the system state $\mathbf{x}$ in LIO systems is represented as an element which evolves on the 18-dimensional compound differentiable manifold $\mathcal{M} \triangleq SO(3) \times {\mathbb{R}^{15}}$ \cite{hertzberg2013integrating, IKFoM-arXiv}: 
    \begin{equation}
    \label{eq:state_manifold}
    \setlength{\arraycolsep}{4.1pt}
    \begin{aligned}
        \mathbf{x} &\triangleq {\begin{bmatrix} \mathbf{R}_{wb}^{\top} & {^{w}\mathbf{p}_{wb}^{\top}} & {^{w}\mathbf{v}_{wb}^{\top}} & {^{b}\mathbf{b}_{g}^{\top}} & {^{b}\mathbf{b}_{a}^{\top}} & {^{w}\mathbf{g}^{\top}} \end{bmatrix}}^{\top} \! \in \! \mathcal{M} \! 
    \end{aligned}
    \end{equation}
Two general operators (``boxplus'': $\boxplus$ and ``boxminus'': $\boxminus$) are defined to operate the state on the tangent space of manifolds \cite{hertzberg2013integrating}. The exponential map $\mathrm{Exp}(\cdot)$ and the logarithmic map $\mathrm{Log}(\cdot)$ on $SO(3)$ \cite{murray2017mathematical_manipulation} are utilized to map elements on the manifold and its tangent space. 

\subsubsection{Problem Description of Initialization} 
Therefore, it is evident from (\ref{eq:state_manifold}) that the objective of initialization modules is to provide relatively accurate initial states and corresponding uncertainties to the subsequent LIO. These initial states include rotation, position, velocity, and gravity with respect to the world frame, as well as IMU biases in the body frame. Additionally, the noise covariance of IMU can also be estimated during the initialization step. 

However, most initialization methods in LIO usually depend on certain motion assumptions. One type is the stationary motion assumption \cite{LINS, FAST-LIO, FAST-LIO2, DLIO, LOAM}, while another is the specific excitation motion assumption \cite{LI-Init}. This reliance stems from the fact that conventional LiDARs can only measure geometric point ranges for online estimators, which results in the immeasurability of velocity and IMU biases. Thus, the initial velocity can be set to zero under the stationary assumption \cite{LINS, FAST-LIO, FAST-LIO2, DLIO, LOAM}, or the velocity state can be estimated using LiDAR-only odometry (LO) during excitation motions \cite{LI-Init}. Nevertheless, owing to the paradox discussed in Section \ref{subsec:crux}, conventional LiDAR measurements are difficult to be utilized in the initialization phase. In contrast, FMCW Doppler LiDARs can directly measure the Doppler velocity of per-return point, which is independent of environmental structures. Thereby, we propose an initialization framework for Doppler LIO systems that leverages inherent Doppler measurements and essential non-inertial kinematic connections, without relying on scan undistortion, motion assumptions, or map correspondences. 
\subsection{Doppler-Inertial Velocimeter} \label{sec:doppler_inertial_velocimeter}
The proposed Doppler-Inertial Velocimeter (DIV) estimates the LiDAR linear velocity, body velocity (both linear and angular), and gyroscope bias using point-wise Doppler velocity and gyroscope data (Fig. (\ref{fig:system_overview})). DIV also outputs pose integration and a static map at an extremely high frequency, ensuring fast convergence and stability of the estimator. The integrated poses and estimated velocities are utilized in the following estimation of accelerometer bias and gravity. 

\subsubsection{State in DIV} 
The state $\mathbf{x}_{V} \! \in \! \mathbb{R}^{12}$ in DIV includes the LiDAR linear velocity $^{l}\mathbf{v}_{wl}$, the body linear velocity $^{b}\mathbf{v}_{wb}$, the body angular velocity $^{b}\bm{\omega}_{wb}$, and the gyroscope bias $^{b}\mathbf{b}_{g}$: 
    \begin{equation*}
    \label{eq:eve_states}
    \begin{aligned}
        \mathbf{x}_{V} &\triangleq {\begin{bmatrix} {^{l}\mathbf{v}_{wl}^{\top}} & {^{b}\mathbf{v}_{wb}^{\top}} & {^{b}\bm{\omega}_{wb}^{\top}} & {^{b}\mathbf{b}_{g}^{\top}} \end{bmatrix}}^{\top} \in \mathbb{R}^{12} \notag 
    \end{aligned}
    \end{equation*}
where the LiDAR velocity is expressed in the LiDAR frame $l$, whereas other states are expressed in the body frame $b$. The error state of DIV can be derived in Euclidean space: 
    \begin{equation*}
    \label{eq:error_state}
    \begin{aligned}
        {\delta} \mathbf{x}_{V} \triangleq \mathbf{x}_{V} \boxminus {\mathbf{x}}_{V_{est}} = {\begin{bmatrix} {^{l}{\delta} \mathbf{v}_{wl}^{\top}} \!\! & \!\! {^{b}{\delta} \mathbf{v}_{wb}^{\top}} \!\! & \!\! {^{b}{\delta} \bm{\omega}_{wb}^{\top}} \!\! & \!\! {^{b}{\delta} \mathbf{b}_{g}^{\top}} \end{bmatrix}}^{\top} \in \mathbb{R}^{12} \notag 
    \end{aligned}
    \end{equation*}
where $\mathbf{x}_{V}$ is the true state and ${\mathbf{x}}_{V_{est}}$ is the nominal estimate. 

\subsubsection{State Propagation}
For the high-frequency point-wise design, the DIV dynamics can be formulated as \cite{stalford1981aerodynamic_identification, savage2000strapdown, xu2022robots_state_estimation}: 
    \begin{equation}
    \label{eq:continuous_model} 
    \begin{aligned}
        {^{l}\dot{\mathbf{v}}_{wl}} \! = \! \mathbf{n}_{v_l}, \! 
        {^{b}\dot{\mathbf{v}}_{wb}} \! = \! \mathbf{n}_{v_b}, \! 
        {^{b}\dot{\bm{\omega}}_{wb}} \! = \! \mathbf{n}_{\omega}, \! 
        {^{b}\dot{\mathbf{b}}_{g}} \! = \! -\frac{1}{\tau_{b_g}} {^{b}\mathbf{b}}_{g} \! + \! \mathbf{n}_{b_g} \! 
    \end{aligned}
    \end{equation}
where $\mathbf{n}_{v_l}$, $\mathbf{n}_{v_b}$, $\mathbf{n}_{\omega}$, and $\mathbf{n}_{b_g}$ are zero-mean Gaussian noises. $\tau_{b_g}$ is the correlation time of the Gauss-Markov process \cite{savage2000strapdown}. To update states when each measurement arrives, we can propagate states with the discrete-time propagation model: 
    \begin{equation*}
    \label{eq:discrete_model}
    \begin{aligned} 
        \mathbf{x}_{V_{i + 1}} &= \mathbf{x}_{V_{i}} \boxplus \left( {\Delta t} \mathbf{f} \left( \mathbf{x}_{V_{i}}, \mathbf{w}_{V_{i}} \right) \right) \notag 
    \end{aligned}
    \end{equation*}
The system noise ${\mathbf{w}_V} \triangleq \begin{bmatrix} \mathbf{n}_{v_l}^{\top} & \mathbf{n}_{v_b}^{\top} & \mathbf{n}_{\omega}^{\top} & \mathbf{n}_{b_g}^{\top} \end{bmatrix}^{\top} \in \mathbb{R}^{12}$, and $\Delta t$ is the time interval between two adjacent time steps. The propagation function $\mathbf{f} \left( \mathbf{x}_{V_{i}}, \mathbf{w}_{V_i} \right) \in \mathbb{R}^{12}$ can be directly derived from (\ref{eq:continuous_model}). Accordingly, the state estimate $\widehat{\mathbf{x}}_{V}$ can be propagated from the last updated state $\overline{\mathbf{x}}_{V_{last}}$: 
    \begin{equation*}
    \label{eq:nominal_propagation}
    \begin{aligned}
        \widehat{\mathbf{x}}_{V_{i + 1}} = \widehat{\mathbf{x}}_{V_{i}} \boxplus \left( {\Delta t} \mathbf{f} \left( \widehat{\mathbf{x}}_{V_{i}}, \mathbf{0}_{12 \times 1} \right) \right), \  \widehat{\mathbf{x}}_{V_{0}} = {\overline{\mathbf{x}}_{V_{last}}} \notag 
    \end{aligned}
    \end{equation*}
The error state and the covariance are propagated as: 
    \begin{equation*}
    \label{eq:cov_propagation}
    \begin{aligned} 
        {{\delta} \mathbf{x}_{V_{i + 1}}} &= \mathbf{F}_{V_{i}} {{\delta} \mathbf{x}_{V_{i}}} + \mathbf{F}_{\mathbf{w}_{i}} {\mathbf{w}_{V_{i}}}, \  {{\delta} \mathbf{x}_{V_{0}}} = {\mathbf{0}_{12 \times 1}} \\ 
        \widehat{\mathbf{P}}_{V_{i + 1}} &= \mathbf{F}_{V_{i}} \widehat{\mathbf{P}}_{V_{i}} \mathbf{F}_{V_{i}}^{\top} + \mathbf{F}_{\mathbf{w}_{i}} \mathbf{Q}_{i} \mathbf{F}_{\mathbf{w}_{i}}^{\top}, \  {\widehat{\mathbf{P}}_{V_{0}}} = {\overline{\mathbf{P}}_{V_{last}}} \notag 
    \end{aligned}
    \end{equation*}
where $\mathbf{Q}_i$ is the noise covariance. $\mathbf{F}_{V_{i}}$ and $\mathbf{F}_{\mathbf{w}_{i}}$ are the constant transition matrix and the noise Jacobian from (\ref{eq:continuous_model}). 

\subsubsection{State Update} 
DIV utilizes point-wise Doppler velocity and gyroscope measurements to update states. Doppler measurements can serve as correspondence-free observations \cite{zhao2024FMCW-LIO, DICP, Need_for_Speed}, eliminating the need for data association from maps or scans during initialization. This approach effectively mitigates the cumulative effects of large initial errors. Moreover, point-wise state updating with individual LiDAR points enables extremely high-frequency outputs, which significantly facilitates the fast convergence of DIV. 

\textbf{Doppler Velocity Observation.} For a LiDAR point $p_j$ sampled at time step $j$, its position measurement with respect to the LiDAR position at time step $j$ is denoted as $^{l_j}{\widetilde{\mathbf{p}}}_{l_j p_j}$. The FMCW Doppler LiDAR can also capture the instant Doppler velocity of per LiDAR point \cite{DICP}. For the static point $p_j$ (i.e., $^{w}\mathbf{v}_{wp_j} \equiv \mathbf{0}$), its Doppler velocity ${\widetilde{v}}^{d}_{j}$ is measured with respect to the LiDAR position and the LiDAR velocity at time step $j$. From the perspective of non-inertial kinematics in (\ref{eq:noninertial_kin_lin_1st}), the true Doppler velocity of $p_j$ is \cite{DICP}: 
    \begin{equation*}
    \label{eq:doppler_j}
    \begin{aligned}
        {{v}^{d}_{j}} = \frac{{^{l_j}{\mathbf{p}}_{l_j p_j}^{\top}}}{\left\| {^{l_j}{\mathbf{p}}_{l_j p_j}} \right\|} {{^{l_j}}\dot{\mathbf{p}}_{l_j p_j}} = - \frac{{^{l_j}{\mathbf{p}}_{l_j p_j}^{\top}}}{\left\| {^{l_j}{\mathbf{p}}_{l_j p_j}} \right\|} {^{l_j}{\mathbf{v}}_{wl_j}} \notag 
    \end{aligned}
    \end{equation*}
where $^{l_j}{\mathbf{v}}_{wl_j}$ is the LiDAR velocity with respect to the world frame and projected in the LiDAR frame at time step $j$. Hence, the observation model ${{{h}}_{d_l} {\left( {\mathbf{x}_{V_j}}, {n_{d_l}} \right)}}$ for LiDAR linear velocity ${^{l}\mathbf{v}_{wl}}$ and the observation model ${{{h}}_{d_b} {\left( {\mathbf{x}_{V_j}}, {n_{d_b}} \right)}}$ for body velocity states can be derived from (\ref{eq:noninertial_kin_lin_1st}): 
    \begin{align}
    \label{eq:doppler_vel_obs}
        {{h}_{d_l} {\left( {\mathbf{x}_{V_j}}, {n_{d_l}} \right)}} \! &\triangleq \! - {^{l_j}{\mathbf{d}}_{l_j p_j}^{\top}} {^{l_j}{\mathbf{v}}_{wl_j}} + {n_{d_l}} \\ 
        {{h}_{d_b} {\left( {\mathbf{x}_{V_j}}, {n_{d_b}} \right)}} \! &\triangleq \! - {^{l_j}{\mathbf{d}}_{l_j p_j}^{\top}} \! \left[ {\mathbf{R}_{bl}^{\top}} \left( {^{b_j}\mathbf{v}_{wb_j}} \!\! + \!\! {^{b_j}{\bm{\omega}}_{wb_j}} \!\! \times \!\! {^{b}\mathbf{p}_{bl}} \right) \right] \! + \! {n_{d_b}} \notag 
    \end{align}
where ${^{l_j}{\mathbf{d}}_{l_j p_j}} \!\! = \!\! \frac{{^{l_j}{\mathbf{p}}_{l_j p_j}}}{\left\| {^{l_j}{\mathbf{p}}_{l_j p_j}} \right\|} \! \in \! \mathbb{R}^{3}$ is the normalized direction vector of point $p_j$. ${n_{d_l}}, {n_{d_b}} \in \mathbb{R}$ are the directly modeled measurement noises. ${\mathbf{R}_{bl}}$, ${^{b}\mathbf{p}_{bl}}$ are LiDAR-IMU extrinsic parameters which can be calibrated offline \cite{LI-Init, lv2020targetless}. When point $p_j$ arrives, the observation residuals ${{r}_{d_{l_j}}}, {{r}_{d_{b_j}}} \in \mathbb{R}$ can be obtained with the propagated state $\widehat{\mathbf{x}}_{V_j}$: 
    \begin{align}
        {{r}_{d_{l_j}}} &= {{h}_{d_l} {\left( {\mathbf{x}_{V_j}}, {{n}_{d_l}} \right)}} - {{h}_{d_l} {\left( {\widehat{\mathbf{x}}_{V_j}}, {0} \right)}} \approx {\mathbf{H}_{d_{l_j}}} {{\delta} \mathbf{x}_{V_j}} + {n_{d_l}} \notag \\ 
        {{r}_{d_{b_j}}} &= {{h}_{d_b} {\left( {\mathbf{x}_{V_j}}, {{n}_{d_b}} \right)}} - {{h}_{d_b} {\left( {\widehat{\mathbf{x}}_{V_j}}, {0} \right)}} \approx {\mathbf{H}_{d_{b_j}}} {{\delta} \mathbf{x}_{V_j}} + {n_{d_b}} \notag 
    \end{align}
where the observation Jacobians ${\mathbf{H}_{d_{l_j}}}, {\mathbf{H}_{d_{b_j}}} \in \mathbb{R}^{1 \times 12}$ are: 
    \begin{align}
    \label{eq:doppler_vel_obs_jac}
        {\mathbf{H}_{d_{l_j}}} &= {\begin{bmatrix} - {^{l_j}{\widetilde{\mathbf{d}}}_{l_j p_j}^{\top}} & {{\mathbf{0}}^{\top}} & {{\mathbf{0}}^{\top}} & {{\mathbf{0}}^{\top}} \end{bmatrix}} \notag \\ 
        {\mathbf{H}_{d_{b_j}}} &= {\begin{bmatrix} {{\mathbf{0}}^{\top}} \! & \! - { ( {\mathbf{R}_{bl}} {^{l_j}{\widetilde{\mathbf{d}}}_{l_j p_j}} ) }^{\top} \! & \! { ( {\mathbf{R}_{bl}} {^{l_j}{\widetilde{\mathbf{d}}}_{l_j p_j}} ) }^{\top} { [ {^{b}\mathbf{p}_{bl}} ] }_{\times} \! & \! {{\mathbf{0}}^{\top}} \end{bmatrix}} \notag 
    \end{align}
To mitigate erroneous observations from dynamic points, points with observation residuals larger than a certain threshold can be efficiently detected and discarded \cite{DICP, zhao2024FMCW-LIO}. 

\textbf{Angular Velocity Observation.} Considering the gyroscope measurement ${^{b_k}{\widetilde{\bm{\omega}}}_k}$ arrives at time step $k$, the angular velocity observation model ${{\mathbf{h}}_{\bm{\omega}} {\left( {\mathbf{x}_{V_k}}, {\mathbf{n}_{g}} \right)}}$ is: 
    \begin{equation}
    \label{eq:ang_vel_obs}
    \begin{aligned}
        {^{b_k}{\widetilde{\bm{\omega}}}_k} &= {{\mathbf{h}}_{\bm{\omega}} {\left( {\mathbf{x}_{V_k}}, {\mathbf{n}_{g}} \right)}} \triangleq {^{b_k}\bm{\omega}_{wb_k}} + {^{b_k}\mathbf{b}_{g}} + {\mathbf{n}_{g}} 
    \end{aligned}
    \end{equation}
where $\mathbf{n}_{g}$ is the gyroscope measurement noise. The observation residual $\mathbf{r}_{{\bm{\omega}}_k} \in \mathbb{R}^{3}$ and the Jacobian ${\mathbf{H}_{{\bm{\omega}}_k}} \in \mathbb{R}^{3 \times 12}$ are: 
    \begin{equation*}
    \label{eq:ang_vel_residual}
    \begin{aligned}
        {\mathbf{r}_{{\bm{\omega}}_k}} = {{\mathbf{h}}_{\bm{\omega}} {\left( {\mathbf{x}_{V_k}}, {\mathbf{n}_{g}} \right)}} - {{\mathbf{h}}_{\bm{\omega}} {\left( {\widehat{\mathbf{x}}_{V_k}}, {\mathbf{0}} \right)}} \approx {\mathbf{H}_{{\bm{\omega}}_k}} {{\delta} \mathbf{x}_{V_k}} + {\mathbf{n}_{g}} \notag 
    \end{aligned}
    \end{equation*}
    \begin{equation*}
    \label{eq:ang_vel_obs_jac}
    \begin{aligned}
        {\mathbf{H}_{{\bm{\omega}}_k}} = \begin{bmatrix} {\mathbf{0}_{3 \times 3}} & {\mathbf{0}_{3 \times 3}} & {\mathbf{I}} & {\mathbf{I}} \end{bmatrix} \notag 
    \end{aligned}
    \end{equation*}

It can be found that the proposed DIV can be formulated as a linear Kalman filter. Therefore, assuming the LiDAR point or the gyroscope measurement arrives at time step $i + 1$, the updated state ${\overline{\mathbf{x}}}_{V_{i + 1}}$ and covariance ${\overline{\mathbf{P}}}_{V_{i + 1}}$ can be obtained from Kalman filtering \cite{bell1993iterated}. For static initialization, zero velocity can be directly detected from (\ref{eq:doppler_vel_obs}) and (\ref{eq:ang_vel_obs}). A strategy similar to the Zero-Velocity Update (ZUPT) \cite{foxlin2005pedestrian_ins_zupt} can be integrated into the framework, yielding a reduced three-dimensional filter specifically for the incremental estimation of gyroscope bias, akin to the mean-based static initialization method \cite{FAST-LIO2, DLIO}. Notably, the DIV filter is fully observable due to the weakly observable gyroscope bias, stemming from the dynamics model in (\ref{eq:continuous_model}). However, as discussed in \cite{Need_for_Speed}, the gyroscope bias is unobservable in the velocity estimator only using the measurements from an FMCW Doppler LiDAR and a gyroscope. Hence, in practice, the gyroscope bias can also be removed from DIV during dynamic initialization and treated as a constant from the prior calibration or the latest update. In this work, the gyroscope bias is retained within DIV to unify the framework for both dynamic and static initialization. 

Thereby, DIV is a high-frequency velocimeter, and the estimated velocities are used for pose integration, static mapping, and the estimation of accelerometer bias and gravity. 

\subsubsection{Pose Integration and Static Mapping} 
Once the velocities are updated at time step $i + 1$, the body pose is integrated from the last update time $i$ under the assumption of fixed-axis rotation \cite{savage2000strapdown}, deriving a trajectory during initialization: 
    \begin{equation}
    \label{eq:pose_integ_continuous}
    \begin{aligned}
        {{\overline{\mathbf{R}}}_{wb_{i + 1}}} &= {{\overline{\mathbf{R}}}_{wb_{i}}} {\mathrm{Exp}} { \left( {\int_{t_{i}}^{t_{i + 1}}} {^{b_{\tau}}{\bm{\omega}}_{wb_{\tau}}} {d \tau} \right) } \\ 
        {^{w}{\overline{\mathbf{p}}}_{wb_{i + 1}}} &= {^{w}{\overline{\mathbf{p}}}_{wb_{i}}} + {\int_{t_{i}}^{t_{i + 1}}} {{\mathbf{R}}_{wb_{\tau}}} {^{b_{\tau}}{\mathbf{v}}_{wb_{\tau}}} {d \tau}
    \end{aligned}
    \end{equation} 
where ${^{b_{\tau}}{\bm{\omega}}_{wb_{\tau}}}$, ${^{b_{\tau}}{\mathbf{v}}_{wb_{\tau}}}$, and ${{\mathbf{R}}_{wb_{\tau}}}$ are the body angular, linear velocity, and rotation in continuous time. The initial pose in the initialization phase ${{\overline{\mathbf{R}}}_{wb_0}} = {\mathbf{I}}$, ${^{w}{\overline{\mathbf{p}}}_{wb_0}} = {\mathbf{0}}$. Since velocities are updated at the arrival of each LiDAR point and gyroscope measurement, the pose integration in (\ref{eq:pose_integ_continuous}) can be computed using accurate higher-order methods \cite{savage2000strapdown}. 

After the pose integration at time step $j$ upon the arrival of LiDAR point $p_j$, the static point $p_j$ can be immediately mapped to the world frame based on the current pose: 
    \begin{equation}
    \label{eq:lidar_point_mapping}
    \begin{aligned}
        {^{w}{\overline{\mathbf{p}}}_{w p_j}} = {{\overline{\mathbf{R}}}_{wb_{j}}} \left( {\mathbf{R}_{bl}} {^{l_j}{\widetilde{\mathbf{p}}}_{l_j p_j}} + {^{b}{\mathbf{p}_{bl}}} \right) + {^{w}{\overline{\mathbf{p}}}_{wb_{j}}} 
    \end{aligned}
    \end{equation} 
where ${^{w}{\overline{\mathbf{p}}}_{w p_j}} \!$ is the updated position of LiDAR point $p_j$ in the world frame. Accordingly, whether the sensor platform is in dynamic or stationary motions during the initialization phase, the proposed framework can achieve accurate pose estimation and static mapping without any motion assumptions. This approach overcomes the paradox in initialization, as discussed in Section \ref{subsec:crux}. In particular, the point-wise scheme eliminates reliance on accumulated LiDAR scans, thereby avoiding low-frequency outputs and the need for additional undistortion or motion compensation methods \cite{FAST-LIO2, LIO-SAM, DLIO, zhao2024FMCW-LIO, LOAM}. 
\subsection{Estimation of Accelerometer Bias and Gravity} 
The estimation of accelerometer bias and gravity can be formulated as an optimization problem based on non-inertial kinematics, utilizing the accumulated data (i.e., LiDAR velocities, body velocities, poses, and accelerometer measurements) during the DIV runtime. The LiDAR acceleration ${^{l_k}{\overline{\mathbf{a}}}_{w l_k}}$ in the LiDAR frame, the body acceleration ${^{w}{\overline{\mathbf{a}}}_{wb_k}}$ in the world frame, the body angular velocity ${^{b_k}{\overline{\bm{\omega}}}_{w b_k}}$, and the body angular acceleration ${^{b_k}{\overline{\dot{\bm{\omega}}}}_{w b_k}}$, can be obtained using numerical differentiation at each IMU measurement time step $k$. The second-order non-inertial kinematics (\ref{eq:noninertial_kin_lin_2nd}) represented by the accelerometer measurement ${^{b_k}{\widetilde{\mathbf{a}}}_k}$ and true states is: 
    \begin{align} 
    \label{eq:acc_gt}
        {\mathbf{R}_{bl}} {{^{l_k}}\mathbf{a}_{w l_k}} &= {\left( {^{b_k}{\widetilde{\mathbf{a}}}_k} - {^{b}\mathbf{b}_a} - {\mathbf{n}_a} \right)} + {\mathbf{R}_{wb_k}^{\top}} {^{w}\mathbf{g}} \notag \\ 
        &\hspace{16pt}  + {\left[ {^{b_k}\bm{\omega}_{w b_k}} \right]_{\times}^{2}} {{^{b}}\mathbf{p}_{bl}} + {\left[ {^{b_k}\dot{\bm{\omega}}_{w b_k}} \right]_{\times}} {{^b}\mathbf{p}_{bl}} \\ 
        {^{w}\mathbf{a}_{wb_k}} &= {\mathbf{R}_{wb_k}} {\left( {^{b_k}{\widetilde{\mathbf{a}}}_k} - {^{b}\mathbf{b}_a} - {\mathbf{n}_a} \right)} + {^{w}\mathbf{g}} 
    \end{align}
where $\mathbf{n}_{a}$ is the accelerometer noise. The accelerometer bias and gravity can be estimated by substituting the above estimates and measurements into the optimization problem: 
    \begin{align}
    \label{eq:acc_bias_gravity_opt}
        \!\!\! {{^{b}{\overline{\mathbf{b}}}_{a}}, \!\! {^{w}{\overline{\mathbf{g}}}}} &= \mathop{\arg\min} \limits_{{^{b}\mathbf{b}_{a}}, {^{w}\mathbf{g}}} {\big(} {\sum_{k}} \big|\big| {^{w}{\overline{\mathbf{a}}}_{wb_k}} \! - \! {{\overline{\mathbf{R}}}_{wb_k}} {\left( {^{b_k}{\widetilde{\mathbf{a}}}_k} \! - \! {^{b}\mathbf{b}_{a}} \right) \! - \! {^{w}\mathbf{g}}} \big|\big| ^{2} \notag \\ 
        &\hspace{28pt} + {\sum_{k}} \big|\big| {\mathbf{R}_{bl}} {{^{l_k}}{\overline{\mathbf{a}}}_{w l_k}} \! - \! {\left( {^{b_k}{\widetilde{\mathbf{a}}}_k} \! - \! {^{b}\mathbf{b}_{a}} \right)} \! - \! {\overline{\mathbf{R}}_{wb_k}^{\top}} {^{w}\mathbf{g}} \notag \\ 
        &\hspace{28pt} - \! {\left[ {^{b_k}{\overline{\bm{\omega}}}_{w b_k}} \right]_{\times}^{2}} {{^{b}}\mathbf{p}_{bl}} \! - \! {\left[ {^{b_k}{\overline{\dot{\bm{\omega}}}}_{w b_k}} \right]_{\times}} {{^b}\mathbf{p}_{bl}} \big|\big| ^{2} {\big)} \notag \\ 
        &\hspace{48pt} \text{s.t.} \  {^{w}\mathbf{g}} \in {\mathbb{S}^2 \left( 9.81 \right)}, {^{b}\mathbf{b}_a} \in {\mathbf{B} {\left( {^{b}\mathbf{b}_a} \right)}} 
    \end{align}
where ${\mathbb{S}^2} \left( r \right)$ is the 2-sphere manifold with range $r$ \cite{hertzberg2013integrating} and ${\mathbf{B} {\left( {^{b}\mathbf{b}_a} \right)}}$ is a plausible set for ${^{b}\mathbf{b}_a}$. To alleviate measurement or difference noises, the above estimates and measurements can be filtered by a low-pass filter \cite{gustafsson1996filtering}. Besides, for more accurate estimates and a more consistent map, the LiDAR bundle adjustment can be applied in the end of initialization with a larger but less efficient optimization \cite{liu2023efficient}. However, in this work, considering the target of fast initialization for LIO systems, the efficient approach is sufficiently accurate for initialization, as well as the subsequent LIO system. 
    \begin{figure}[!t]
       \centering
       \includegraphics[width=0.47\textwidth]{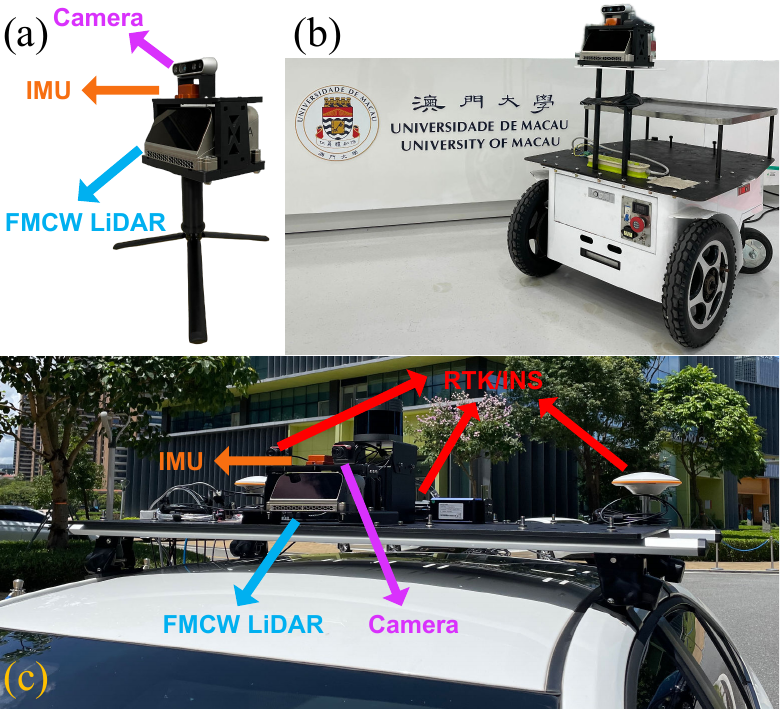}
       \caption{The FMCW Doppler LiDAR-inertial sensor suite can be mounted on (a) handheld, (b) wheeled, and (c) vehicular platforms.} 
       \label{fig:sensor_suite}
    \end{figure}
    \begin{table}[!t] 
        \setlength
        \tabcolsep{4.8pt}
        \caption{Description of Data Sequences} 
        \label{tab:table_data_distance} 
        \begin{center} 
            \begin{tabular}{ccccc} 
                \toprule 
                Sequence & Platform & Motion Feature & Initial Velocity & Distance (m) \\ 
                \midrule
                \textit{dyna\_01} & Handheld & Rotational & 5 rad/s & 70 \\ 
                \textit{dyna\_02} & Handheld & Rotational & 5 rad/s & 75 \\ 
                \textit{dyna\_03} & Handheld & Translational & 5 m/s & 103 \\ 
                \textit{dyna\_04} & Handheld & Rotational & 10 rad/s & 311 \\ 
                \textit{dyna\_05} & Vehicular & Translational & 75 km/h & 100 \\ 
                \textit{dyna\_06} & Vehicular & Translational & 60 km/h & 100 \\ 
                \midrule 
                \textit{stat\_01} & Handheld & Stationary & 0 m/s, 0 rad/s & 72 \\ 
                \textit{stat\_02} & Wheeled & Stationary & 0 m/s, 0 rad/s & 123 \\ 
                \textit{stat\_03} & Wheeled & Stationary & 0 m/s, 0 rad/s & 130 \\ 
                \textit{stat\_04} & Handheld & Stationary & 0 m/s, 0 rad/s & 314 \\ 
                \bottomrule
            \end{tabular}
        \end{center}
    \end{table}
    \begin{table*}[t!]
        \setlength
        \tabcolsep{5.3pt}
        \caption{Absolute Translational Errors (RMSE, m) and End-to-End Errors (End to End, m) of Localization Results} 
        \label{tab:table_localization_results} 
        \begin{center}
            \begin{tabular}{c|c|c|cccccc|cccc} 
                \hline 
                System & Metric & Method & \textit{dyna\_01} & \textit{dyna\_02} & \textit{dyna\_03} & \textit{dyna\_04} & \textit{dyna\_05} & \textit{dyna\_06} & \textit{stat\_01} & \textit{stat\_02} & \textit{stat\_03} & \textit{stat\_04} \\ 
                \hline 
                \multirow{4}{*}{FAST-LIO2 \cite{FAST-LIO2}} & \multirow{2}{*}{RMSE} & Default & 8.21 & $\times$ & 1.92 & $\times$ & 2.55  & 28.67 & 0.20 & 0.36 & \textbf{0.10} & 0.31 \\ 
                                           & & Free-Init & \textbf{0.25} & \textbf{0.10} & \textbf{0.09} &\textbf{0.15} & \textbf{0.92}  &  \textbf{2.02} & \textbf{0.18} & \textbf{0.11} & \textbf{0.10} & \textbf{0.04} \\ 
                \cline{2-13}
                & \multirow{2}{*}{End to End} & Default & 6.65 & $\times$ & 2.81 & $\times$ & $-$ & $-$ & 0.36 & 0.25 & 0.24 & \textbf{0.08} \\
                                           & & Free-Init & \textbf{0.51} & \textbf{0.13} & \textbf{0.25} &\textbf{0.58} & $-$ & $-$ & \textbf{0.21} & \textbf{0.16} & \textbf{0.22} & \textbf{0.08} \\ 
                \hline 
                \hline 
                \multirow{4}{*}{DLIO \cite{DLIO}} & \multirow{2}{*}{RMSE} & Default & 5.27 & 6.23 & $\times$ & 2.01 & 36.90 & 42.83 & 1.17 & 0.91 & 1.29 & 0.57 \\ 
                                           & & Free-Init & \textbf{0.38} & \textbf{0.33} & \textbf{0.45} &\textbf{0.39} & \textbf{1.10} & \textbf{3.88} & \textbf{0.42} & \textbf{0.30} & \textbf{0.69} & \textbf{0.42} \\ 
                \cline{2-13} 
                & \multirow{2}{*}{End to End} & Default & 5.34 & 11.98 & $\times$ & 5.02 & $-$ & $-$ & 1.01 & 1.93 & \textbf{0.03} & 1.37 \\ 
                                           & & Free-Init & \textbf{0.64} & \textbf{0.56} & \textbf{1.08} &\textbf{1.99} & $-$ & $-$ & \textbf{0.10} & \textbf{0.12} & 0.04 & \textbf{1.16} \\ 
                \hline 
            \end{tabular}
        \end{center}
            \footnotesize{
                $\times$ The cross denotes that the system fails or severely drifts in the corresponding sequence. \\
                $-$ The trajectories in the corresponding two sequences (\textit{dyna\_05} and \textit{dyna\_06}) are not end-to-end. 
            }
    \end{table*} 
\subsection{Initial State for Subsequent LIO Systems} 
After the above optimization, let $k$ denote the last time step in initialization. The initial state $\mathbf{x}_{0}$ for the subsequent LIO system (\ref{eq:state_manifold}) is assigned with the estimated states as: 
    \begin{equation*}
    \label{eq:initial_state_LIO}
    \setlength{\arraycolsep}{2.8pt}
    \begin{aligned}
        {\mathbf{x}_{0}} = {\begin{bmatrix} {\overline{\mathbf{R}}_{wb_k}^{\top}} &  {^{w}\overline{\mathbf{p}}_{wb_k}^{\top}} & \left( {\overline{\mathbf{R}}_{wb_k}} {^{b_k}\overline{\mathbf{v}}_{wb_k}} \right)^{\top} & {^{b_k}\overline{\mathbf{b}}_{g}^{\top}} & {^{b}\overline{\mathbf{b}}_{a}^{\top}} & {^{w}\overline{\mathbf{g}}^{\top}} \end{bmatrix}}^{\top} \notag 
    \end{aligned}
    \end{equation*}
where $^{b}\overline{\mathbf{b}}_{a}$ and $^{w}\overline{\mathbf{g}}$ are solved from (\ref{eq:acc_bias_gravity_opt}). The velocity and gyroscope bias states are updated from DIV. The body pose in the world frame is integrated by (\ref{eq:pose_integ_continuous}) up to the last time step $k$. Ultimately, as depicted in Fig. \ref{fig:system_overview}, the estimates and the static map from Free-Init are fed into the subsequent LIO, facilitating rapid convergence, stability, and robustness. 


\section{Experiments} \label{sec:experiments}

\subsection{Experiment Design and Data Collection} 
To assess the effectiveness of the proposed initialization method, we perform comprehensive experiments and analyze its performance based on both quantitative and qualitative results. Considering the short duration of initialization and different initialization methods can lead to varying levels of localization accuracy and mapping consistency in the subsequent LIO, we integrate Free-Init into two state-of-the-art LIO systems: FAST-LIO2 \cite{FAST-LIO2} (tightly-coupled) and DLIO \cite{DLIO} (loosely-coupled), while maintaining identical parameters and configurations as the default version, except for the initialization module. We evaluate localization accuracy (Section \ref{sec:experiments_localization}), mapping consistency (Section \ref{sec:experiments_mapping}), initialization performance in structure-degenerated environments (Section \ref{sec:experiments_degenerated}), as well as velocity estimation accuracy and time consumption (Section \ref{sec:experiments_vel_time}) across diverse platforms, scenarios, and initial motion features. 

    \begin{figure}[!t]
       \centering
       \includegraphics[width=0.47\textwidth]{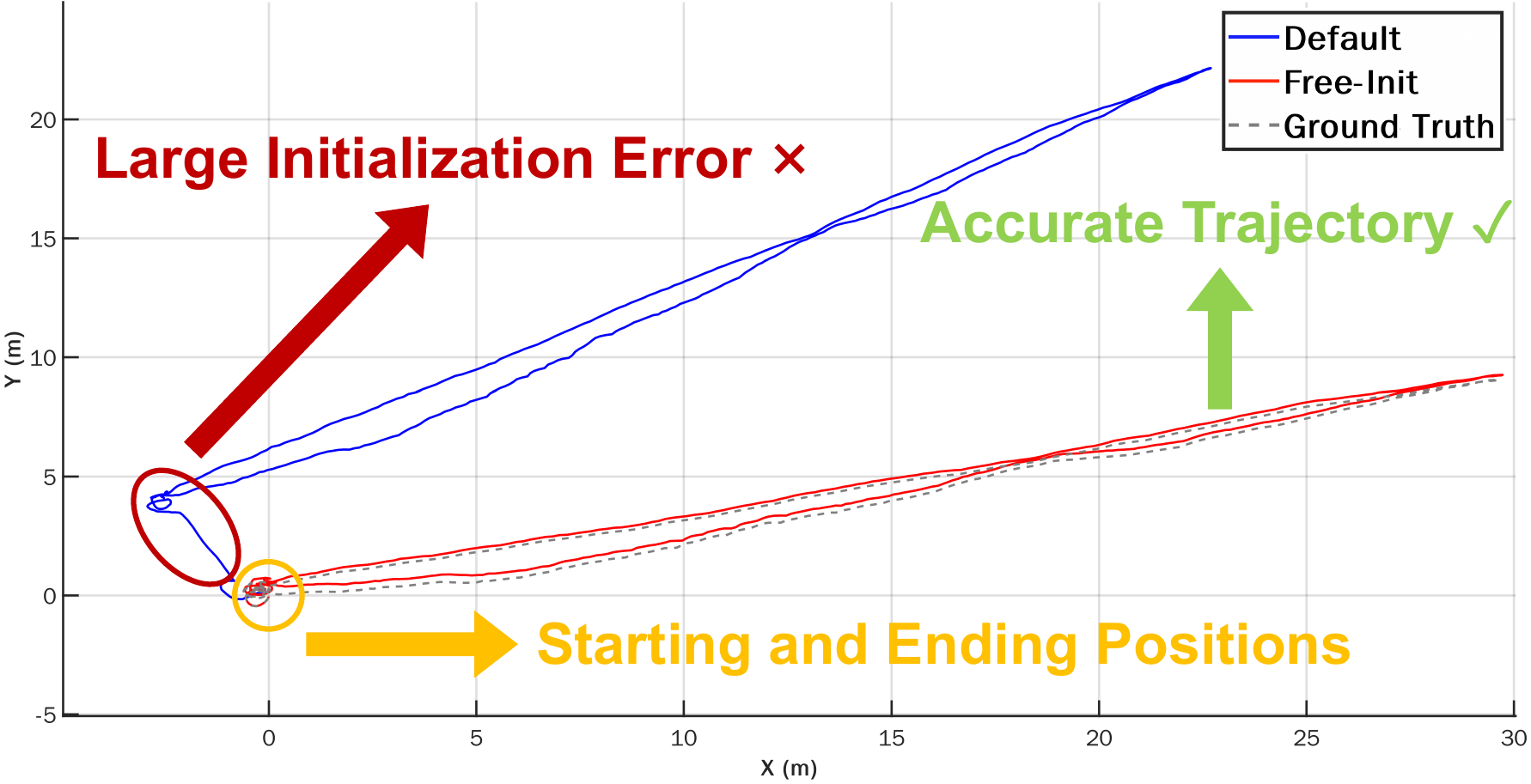}
       \caption{Comparison of trajectory estimation on \textit{dyna\_01}.} 
       \label{fig:traj}
    \end{figure}

    \begin{figure}[!t]
       \centering
       \includegraphics[width=0.47\textwidth]{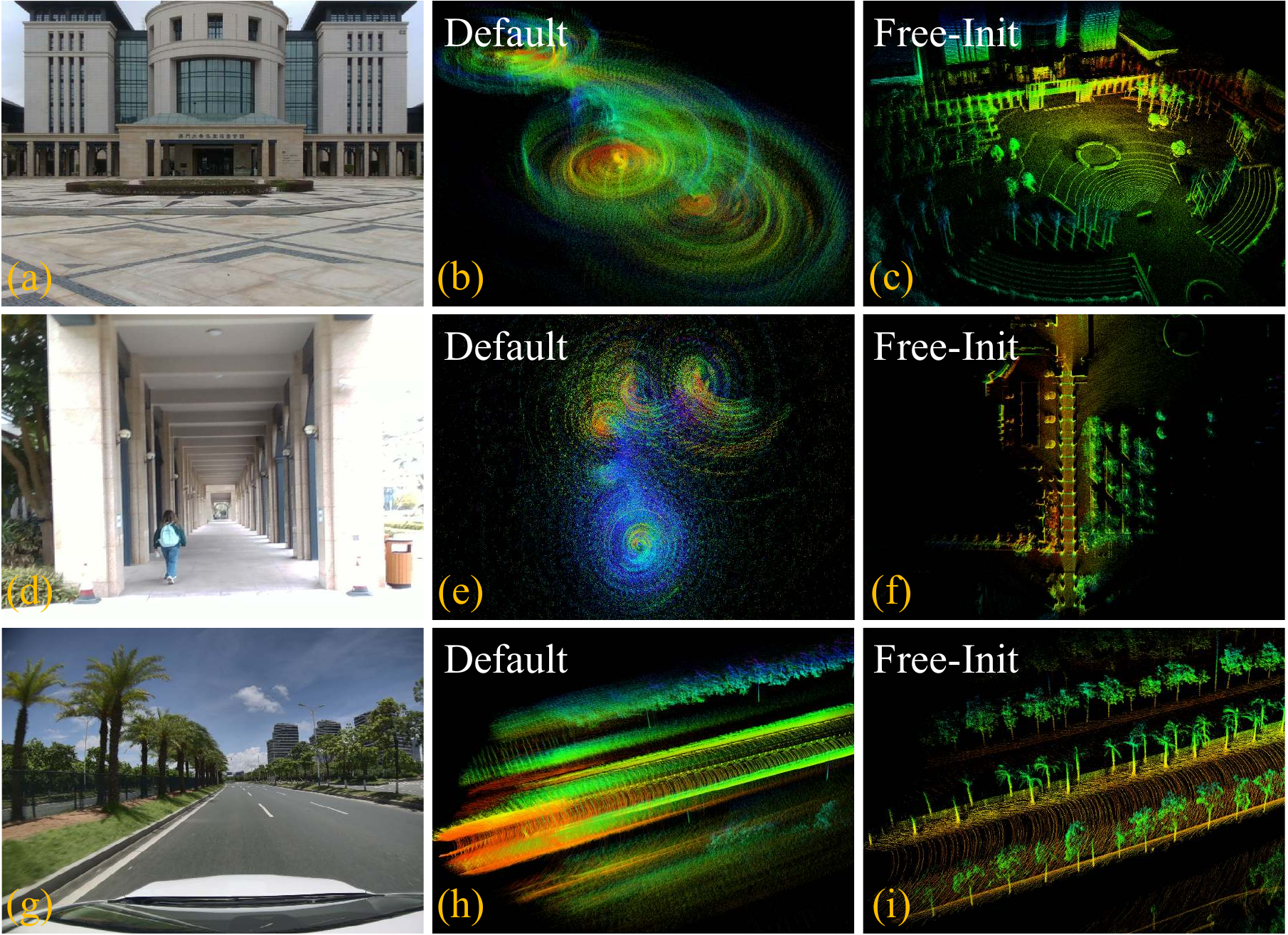}
       \caption{Mapping comparison with the default initialization and Free-Init. Left column: FPV camera images. Middle column: Mapping results with the default initialization in the initial period. Right column: Mapping results with Free-Init in the initial period.} 
       \label{fig:mapping_consistency}
    \end{figure}

In \cite{HeLiPR}, the authors release a heterogeneous LiDAR dataset (HeLiPR) for place recognition tasks, including an FMCW Doppler LiDAR. However, the velocity of LiDAR points in HeLiPR is the absolute velocity after post processing, rather than the raw Doppler velocity. This limitation renders HeLiPR unsuitable for online state estimation. Thus, given the absence of public FMCW Doppler LiDAR-inertial datasets containing raw Doppler measurements, it is necessary to collect a dataset to validate the effectiveness of the proposed framework. An FMCW Doppler LiDAR, Aeva Aeries II \cite{aeva2023}, and an Xsens MTi-G-710 IMU are used to assemble our FMCW Doppler LiDAR-inertial sensor suite. The sensor suite can be mounted on handheld, wheeled, and vehicular platforms (Fig. \ref{fig:sensor_suite}). The frequency of IMU measurements is 200 Hz, and the camera on the sensor suite is only for First-Person-View (FPV) recording. The computing device is equipped with an Intel i7-1165G7 CPU. The sensor suite is held by a person on the handheld platform. The speed of the handheld platform is slow (about 1.5 m/s), while the wheeled platform moves faster (up to 6.5 m/s). The speed of the vehicular platform is significantly higher, ranging from 60 to 75 km/h. It is also ensured that the starting and ending positions of data sequences (except for \textit{dyna\_05} and \textit{dyna\_06}) precisely coincide to facilitate the end-to-end evaluation of localization accuracy. The RTK GNSS provides ground-truth positions for the handheld and wheeled platforms. For the vehicular platform, the dual RTK GNSS integrated with inertial navigation systems (RTK/INS) offers 6-DoF pose and velocity truths (Fig. \ref{fig:sensor_suite}(c)). To validate the effectiveness of Free-Init for dynamic initialization, the \textit{dyna} sequences in Table \ref{tab:table_data_distance} commence with vigorous dynamic rotational and translational motions, while the \textit{stat} sequences remain stationary for approximately the initial 1 to 2 seconds. The distances and initial motion features of data sequences are presented in Table \ref{tab:table_data_distance}. In Free-Init, the data accumulation time is a configurable parameter, and in the experiments, we maintain consistent parameters and settings across the compared methods. The demonstration video of experiments is available at: \href{https://youtu.be/FbyzvJ-4bHI}{https://youtu.be/FbyzvJ-4bHI}.
\subsection{Localization Accuracy} \label{sec:experiments_localization}
We embed Free-Init into two LIO systems and evaluate its impact on localization accuracy. Notably, we maintain identical system parameters when evaluating different methods, only except for the initialization module. The results are shown in Table \ref{tab:table_localization_results}. In \textit{dyna} sequences, the system starts with severely dynamic rotational and translational motions. It can be found that in the \textit{dyna} sequences, Free-Init significantly outperforms the default initialization method on localization accuracy. Particularly, in \textit{dyna\_02} and \textit{dyna\_04} sequences where the system starts with both fast translational and rotational motions, the default method leads to system failures of the subsequent LIO. Contrarily, the system equipped with Free-Init exhibits high accuracy in returning to starting positions, thanks to precise estimation of velocities and poses during initialization. The estimated trajectories of different methods with FAST-LIO2 \cite{FAST-LIO2} in \textit{dyna\_01} are depicted in Fig. \ref{fig:traj}, and the default method yields considerable errors (6.65 m) due to the dynamic motion during initialization. Notably, the fast vehicle speeds in \textit{dyna\_05} and \textit{dyna\_05} pose significant challenges for both initialization and subsequent LIO systems. However, systems with Free-Init can still achieve accurate localization. In \textit{stat} sequences, Free-Init achieves comparable yet more accurate localization accuracy than the default initialization. This improvement arises from the inherent dynamic point removal and the ZUPT-like strategy during static initialization, enabling the establishment of a static map. This static map can further provide consistent constraints for the subsequent LIO and prevent potential drifts. In contrast, the default methods discard all LiDAR points during initialization, thereby failing to provide sufficient constraints in the initial running stage of the subsequent LIO. The results reveal that a resilient initialization module can offer accurate initial states for online systems and prevent failures in challenging dynamic cases, regardless of loosely-coupled or tightly-coupled frameworks. 
    \begin{figure}[!t]
       \centering
       \includegraphics[width=0.47\textwidth]{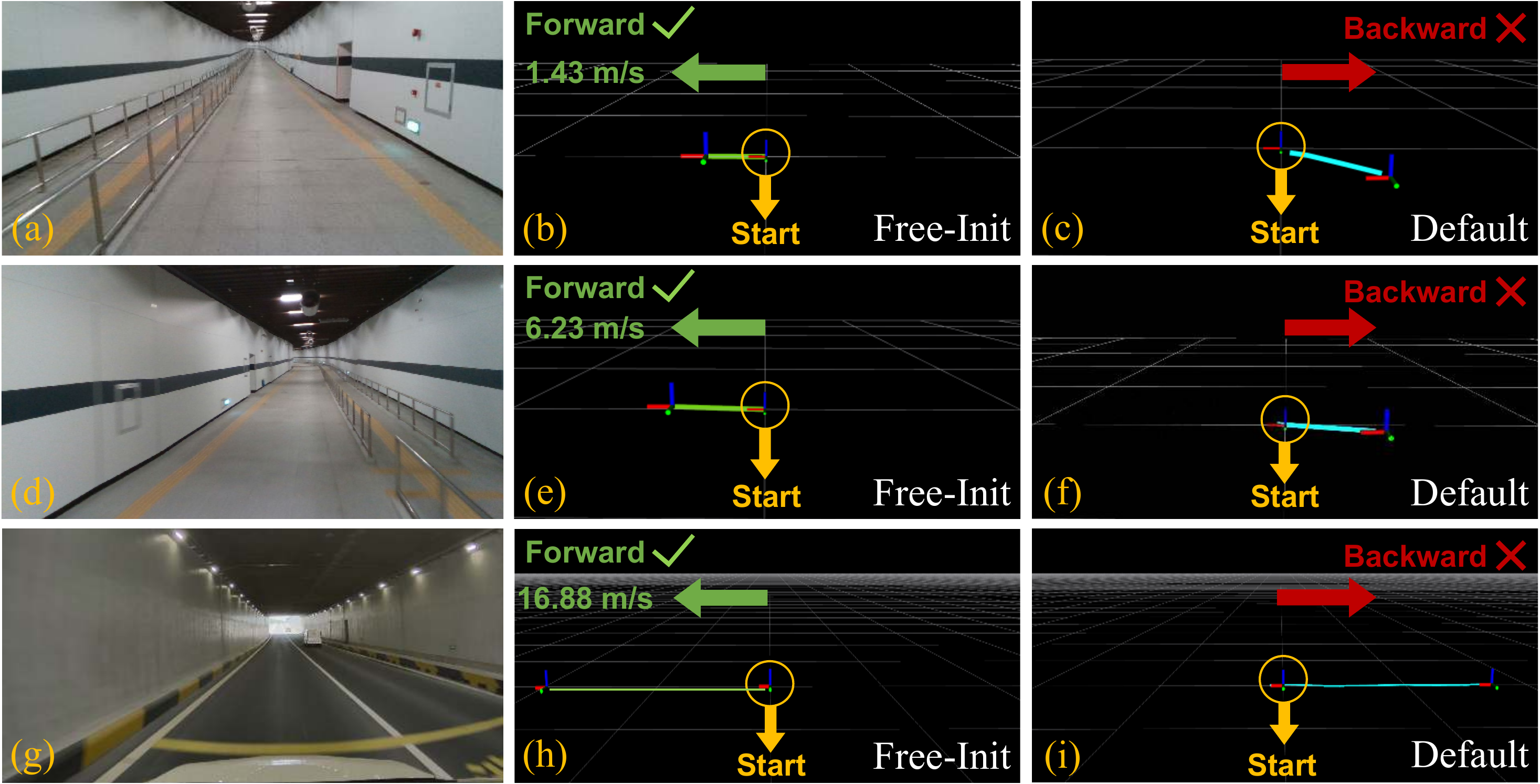}
       \caption{Initialization in degenerate tunnels on the forward moving handheld (top), wheeled (middle), and vehicular (bottom) platforms. Left column: FPV camera images. Middle column: Accurate forward velocity estimation with Free-Init. Right column: Erroneous backward motion estimation with the default initialization.} 
       \label{fig:init_degenerated} 
    \end{figure}
\subsection{Mapping Consistency} \label{sec:experiments_mapping} 
After the initialization and a few seconds of subsequent LIO running, we can evaluate the estimation accuracy and consistency of different initialization methods from the established LiDAR maps. The dynamic initial motions render system failures with the default initialization, outputting chaotic maps. Contrarily, systems with Free-Init can handle these violent initial motions, generating distinct and consistent maps. In the top row of Fig. \ref{fig:mapping_consistency}, the sensor suite is handheld by a person who runs vigorously, performing intense rotational motions in the front of the building. Simultaneously, the system initiates with fast shaking and intense rotations during the person's running. It can be found that the system with the default initialization collapses and outputs an extremely chaotic map (Fig. \ref{fig:mapping_consistency}(b)). On the contrary, the system with Free-Init can precisely estimate initial states while outputs a consistent, neat, and accurate map of the environment (Fig. \ref{fig:mapping_consistency}(c)). Similarly, the middle row of Fig. \ref{fig:mapping_consistency} shows the mapping results during initialization in a corridor with rapid and substantial rotational motions. As expected, the system with the default initialization suffers from a system crash, deriving a messy and meaningless map (Fig. \ref{fig:mapping_consistency}(e)). While the system with Free-Init can construct a sharp map, portraying distinct structures of the corridor. The top view of the corridor map is shown in Fig. \ref{fig:mapping_consistency}(f). As shown in Fig. \ref{fig:mapping_demo} and the bottom row of Fig. \ref{fig:mapping_consistency}, the sensor suite is mounted on both the wheeled platform with a forward speed of 6.5 m/s and the vehicular platform with a forward speed of 75 km/h. The system initiates when the platforms are moving forward on the road. Obviously, the high speeds lead to vague mapping outputs even system collapses with the default initialization. However, the LiDAR maps generated by systems with Free-Init are consistent and clear, enabling them to offer accurate LiDAR correspondences for subsequent LIO systems. 
    \begin{figure}[!t]
       \centering
       \includegraphics[width=0.47\textwidth]{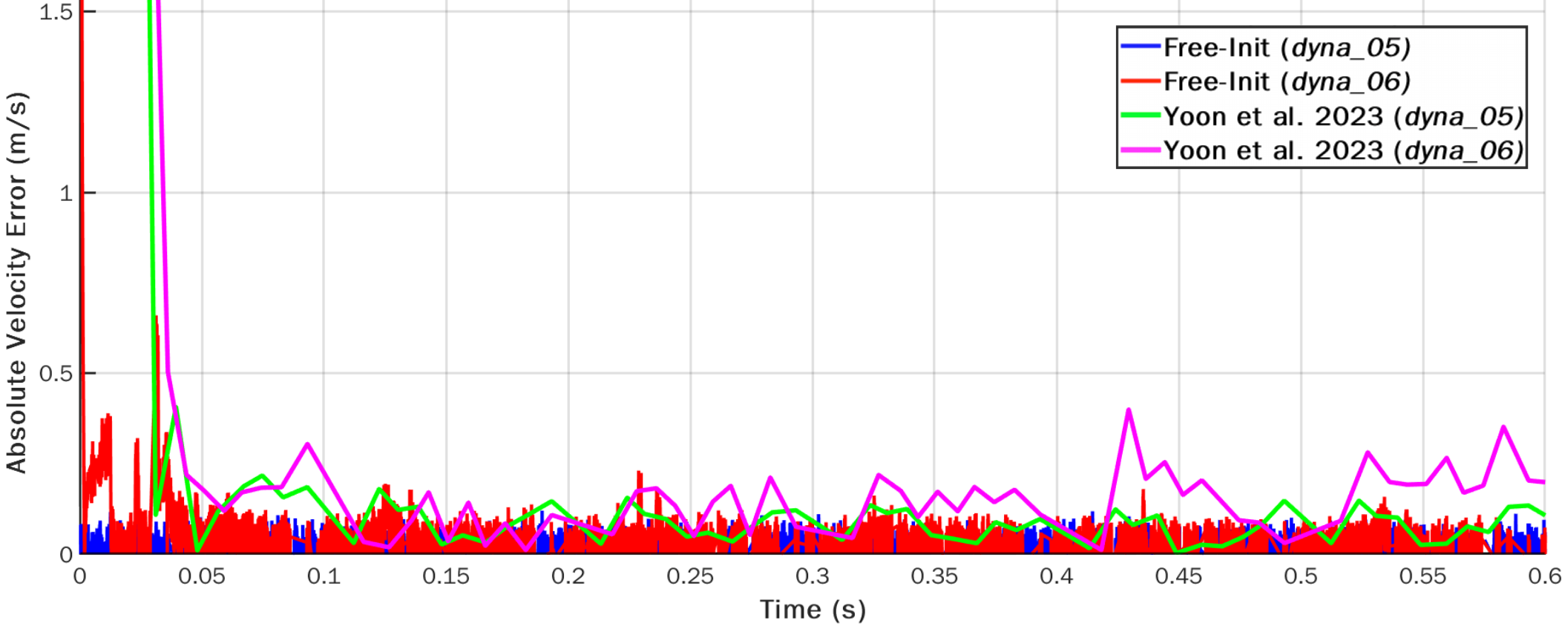}
       \caption{Comparison of velocity estimation on \textit{dyna\_05} and \textit{dyna\_06}.} 
       \label{fig:vel_esti}
    \end{figure}
    \begin{table}[t!] 
        \caption{Average Running Time (s) of Whole Framework} 
        \label{tab:table_avg_running_time} 
        \begin{center} 
            \begin{tabular}{c|c|c} 
                \hline 
                Sequence Type & \textit{dyna} Sequences & \textit{stat} Sequences \\ 
                \hline 
                Average Running Time & 0.208 & 0.116 \\ 
                \hline 
            \end{tabular}
        \end{center}
    \end{table}
\subsection{Initialization in Structure-Degenerated Environments} \label{sec:experiments_degenerated} 
An insight from the correspondence-free observation \cite{DICP, Need_for_Speed, zhao2024FMCW-LIO} in the proposed framework is that the observations are independent of environmental structures, even in structure-degenerated scenes. As a result, Free-Init is expected to offer accurate initial states for online LIO in both structured and structure-degenerated environments, regardless of dynamic or stationary motions. On the contrary, typical LIO systems estimate system states solely based on geometric observations, which often leads to estimation failures in degenerate cases. In particular, in scenarios where the LIO system is moving within a degenerate environment, Free-Init should be capable of accurately estimating the initial velocity for subsequent LIO. As shown in Fig. \ref{fig:init_degenerated}, we test different initialization methods in degenerate tunnels on all forward moving handheld, wheeled, and vehicular platforms. It can be found that Free-Init can provide accurate forward velocity estimates to subsequent LIO systems (about 1.43 m/s, 6.23 m/s, and 16.88 m/s on the handheld, wheeled, and vehicular platforms, respectively), even in these degenerate scenes. However, due to the inability of the default initialization to estimate accurate initial velocity states and the lack of geometric features, the system exhibits erroneous backward motions immediately upon startup, ultimately leading to system failures. The results reveal that the Doppler-aided and correspondence-free observations in the proposed method are remarkably effective, especially in degenerate scenarios. The test results in degenerate scenarios also indicate that the proposed framework overcomes the geometric-correspondence-dependency paradox inherent in the initialization of conventional LIO systems (discussed in Section \ref{subsec:crux}), fundamentally resolving the issue from the standpoint of the correspondence-free first-order kinematics. 
\subsection{Velocity Estimation and Time Consumption} \label{sec:experiments_vel_time}
We also evaluate the velocity estimation accuracy of the proposed Doppler-Inertial Velocimeter (DIV) and the method in Yoon et al. 2023 \cite{Need_for_Speed} on \textit{dyna\_05} and \textit{dyna\_06}, leveraging the high-frequency ground truth from RTK/INS. The absolute velocity error results are shown in Fig. \ref{fig:vel_esti}. Under the point-wise updating scheme, DIV achieves accurate velocity estimation with an extremely high output frequency exceeding 10 kHz. The point-wise updating design eliminates the reliance on explicit motion compensations for both range and Doppler measurements \cite{zhao2024FMCW-LIO}, as well as the need for specific interpolation models. Additionally, the average running time of the entire framework of Free-Init is presented in Table \ref{tab:table_avg_running_time}.


\section{Conclusion} 
This letter presents Free-Init, a LiDAR scan-free, excitation motion-free, and map correspondence-free initialization framework for Doppler LIO systems. The design of Free-Init is enhanced by point-wise Doppler velocity, aligning consistently with the sensing nature of Doppler LiDARs. The primary focus of this work is an initialization framework that aims to achieve a near-ideal initialization design without relying on motion or observation assumptions. This is also the first work to comprehensively showcase the performance potential of Doppler LiDARs across varied platforms, including handheld, robotic, and vehicular platforms. Furthermore, a pivotal insight in this work is that the core solution for initialization lies in the first-order tangent space of system dynamics (e.g., the first-order kinematics of LIO). More importantly, beyond the initialization of Doppler LIO systems, this framework can also be applied to offline multi-session mapping, robot kidnapping cases, online re-initialization, and online startup or switching of state estimators in challenging scenarios. In addition, the Doppler-inertial velocimeter can provide high-frequency velocity estimates for online planners and controllers, facilitating agile and aggressive maneuvers. Overall, the Doppler-aided methodology highlights the potential of Doppler LiDARs for future research and real-world applications, unveiling the essential role of the intrinsically velocity-aware sensors in robotic sensing and estimation systems. 


\bibliographystyle{IEEEtran}
\IEEEtriggeratref{24}
\bibliography{ref}

\end{document}